\documentclass[runningheads]{llncs}

\usepackage{eccv}

\usepackage{eccvabbrv}

\usepackage{graphicx}
\usepackage{booktabs}

\usepackage[accsupp]{axessibility}  

\usepackage{hyperref}

\usepackage{orcidlink}
\usepackage{multirow}
\usepackage{wrapfig}

\begin{document}
\newcommand{\CCH}[1]{{\color{black}#1}\normalfont} 
\newcommand{\MC}[1]{{\color{black}#1}\normalfont} 
\title{DetailSemNet: Elevating Signature Verification through Detail-Semantic Integration}

\titlerunning{Elevating Signature Verification through Detail-Semantic Integration}

\author{
Meng-Cheng Shih\inst{1}\orcidlink{0009-0007-4147-6960} \and
Tsai-Ling Huang\inst{1} \and
Yu-Heng Shih\inst{1} \and
Hong-Han Shuai\inst{1}\orcidlink{0000-0003-2216-077X} \and
Hsuan-Tung Liu\inst{2} \and
Yi-Ren Yeh\inst{3} \and
Ching-Chun Huang\inst{1}\orcidlink{0000-0002-4382-5083}
}

\authorrunning{Shih et al.}

\institute{National Yang Ming Chiao Tung University, Taiwan \email{\{mcshih.ee11,christina.ii12,ra890927.cs12,hhshuai,chingchun\}@nycu.edu.tw} \and
E.SUN Financial Holding Co., Ltd, Taiwan \email{ahare-18342@esunbank.com.tw}\and
National Kaohsiung Normal University, Taiwan
\email{yryeh@nknu.edu.tw}
\url{https://github.com/nycu-acm/DetailSemNet_OSV}}

\maketitle
\begin{abstract}
\CCH{Offline signature verification (OSV) is a frequently utilized technology in forensics. This paper proposes a new model, \textbf{DetailSemNet}, for OSV. Unlike previous methods that rely on holistic features for pair comparisons, our approach underscores the significance of fine-grained differences for robust OSV. We propose to match local structures between two signature images, significantly boosting verification accuracy. Furthermore, we observe that without specific architectural modifications, transformer-based backbones might naturally obscure local details, adversely impacting OSV performance. To address this, we introduce a \textbf{Detail-Semantics Integrator}, leveraging feature disentanglement and re-entanglement. This integrator is specifically designed to enhance intricate details while simultaneously expanding discriminative semantics, thereby augmenting the efficacy of local structural matching. We evaluate our method against leading benchmarks in offline signature verification. Our model consistently outperforms recent methods, achieving state-of-the-art results with clear margins. The emphasis on local structure matching not only improves performance but also enhances the model's interpretability, supporting our findings. Additionally, our model demonstrates remarkable generalization capabilities in cross-dataset testing scenarios. The combination of generalizability and interpretability significantly bolsters the potential of \textbf{DetailSemNet} for real-world applications. }
\end{abstract}


\section{Introduction}
\label{sec:intro}
\CCH{
\begin{figure}[t]
\centering
\begin{minipage}[c]{0.5\linewidth} 
\includegraphics[width=\linewidth]{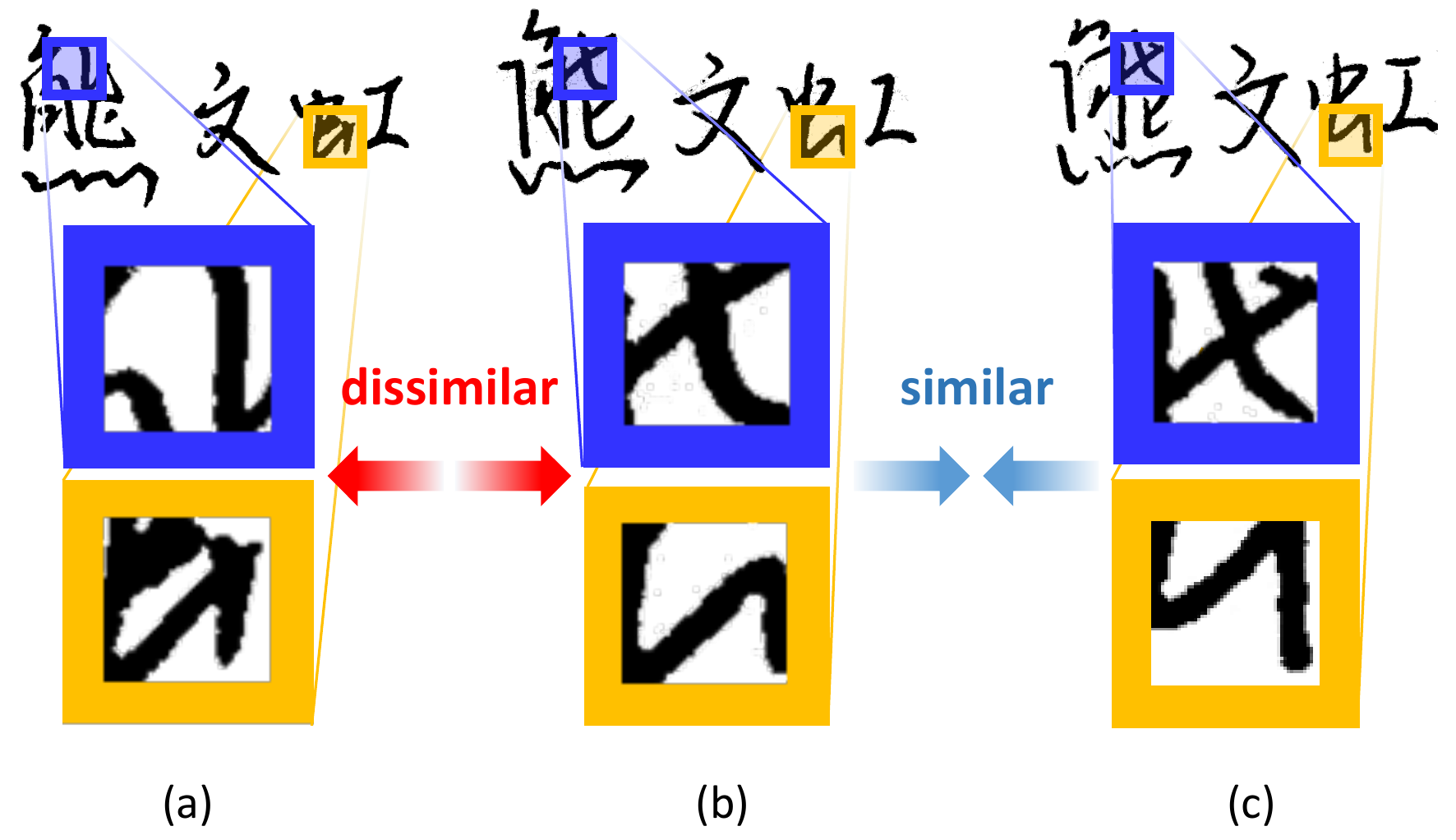}
\caption{Three samples from the ChiSig dataset. Signature (a) originates from a different individual than signatures (b) and (c). At first glance, these signatures appear remarkably similar when viewed holistically. However, detailed analysis at the patch level reveals distinct differences between them, which are aspects frequently overlooked in previous methodologies.}
\label{fig_intro}
\end{minipage}
\begin{minipage}[c]{0.49\linewidth} 
\includegraphics[width=\linewidth]{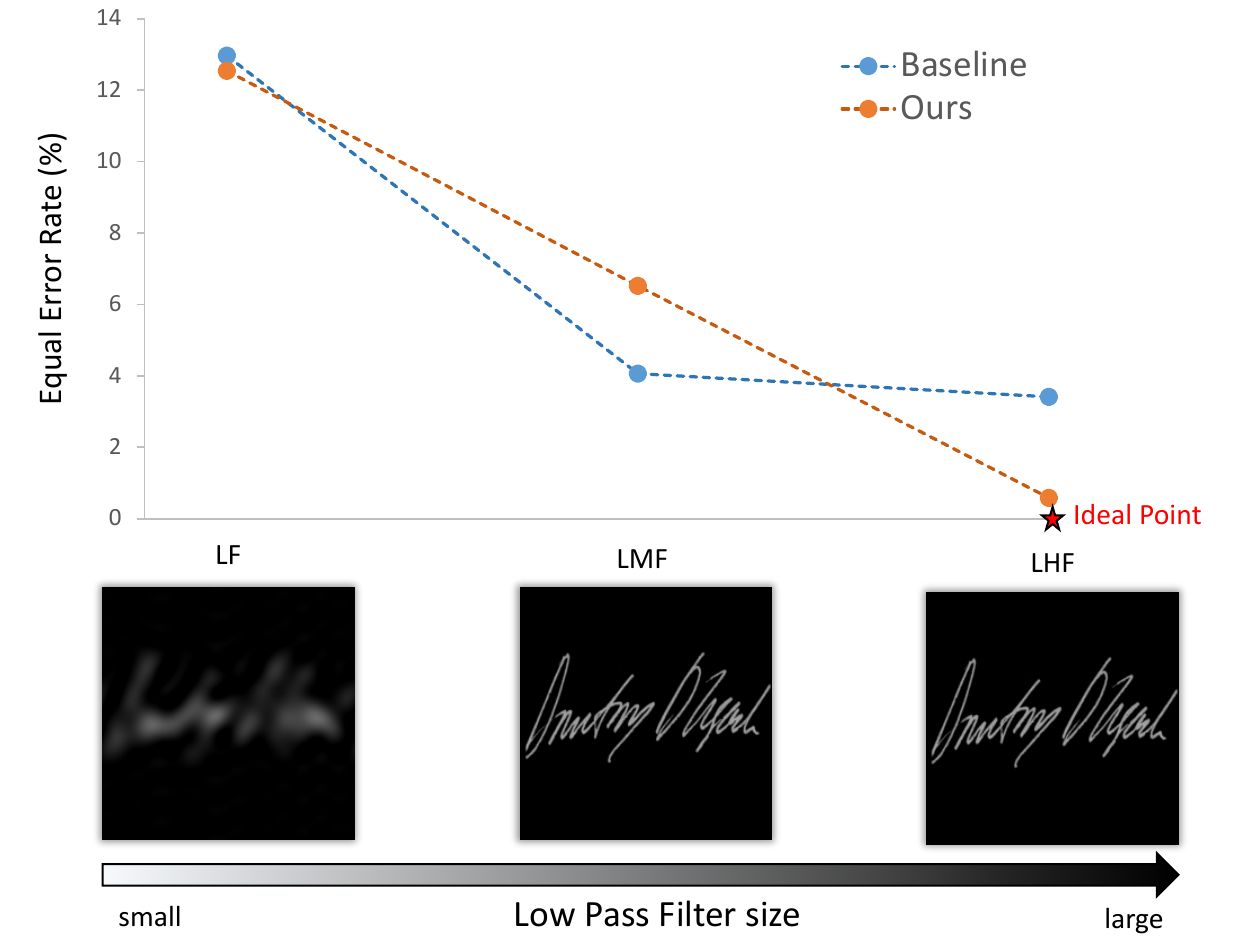}
\caption{We employ filters to extract Low-frequency (LF), low-plus-middle frequency (LMF), and low-plus-high frequency (LHF) images. Our model captures both semantic pattern (low-frequency) and stroke structure and style detail (high-frequency) for improved verification. Leveraging high-frequency data enhances performance, unlike the baseline transformer model, which solely relies on low-frequency patterns and does not benefit from high-frequency features.}
\label{fig_intro_fft}
\end{minipage}
\end{figure}

\begin{figure}[t]
\centering
\includegraphics[width=\textwidth]{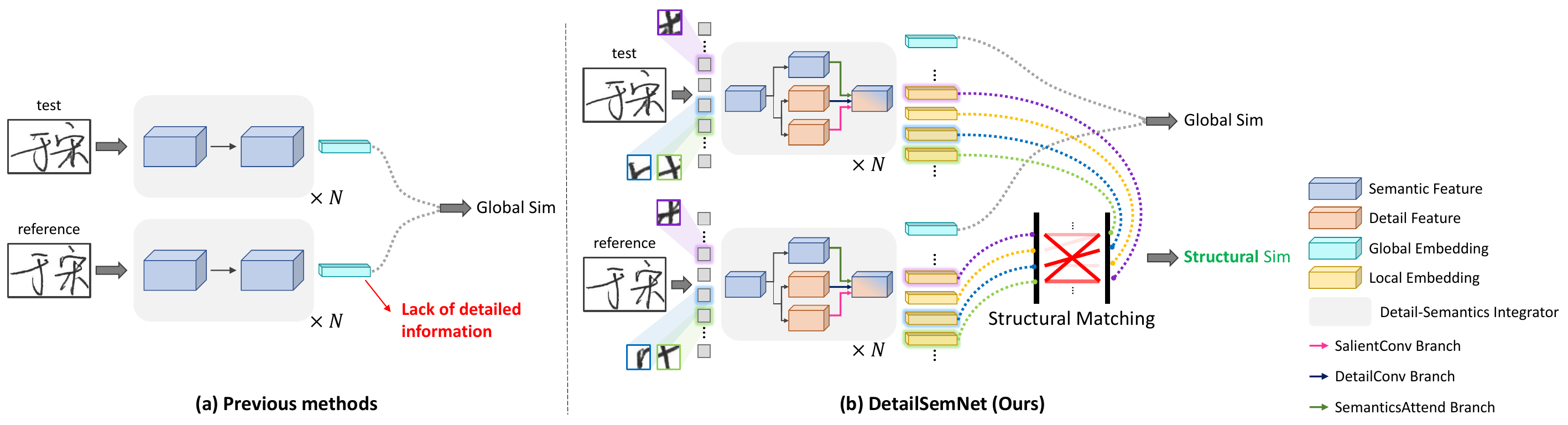}
\caption{Conventional OSV method vs. Our proposed method: The left figure shows the traditional approach, lacking detailed feature information and relying solely on global similarity for comparison. On the right, our method, called \textbf{DetailSemNet}, employs the \textbf{Detail-Semantics Integrator} to divide features into Semantic and Detail components. The Semantic component acquires contextual information through the SemanticsAttend Branch, while the Detail component is processed via the SalientConv and DetailConv Branches. Integrating these outputs yields feature representations containing both detailed and semantic information. Additionally, the model utilizes Structural Matching techniques to emphasize detailed information alongside global similarity.}
\label{fig_compare}
\end{figure}
}

\CCH{
Handwritten offline signature verification (OSV) is a pivotal biometric technology, especially in sectors like banking and commerce. The core goal of this technology is to authenticate a signature by comparing it against a known original. This comparison involves analyzing a test signature alongside a reference signature, allowing the system to determine whether the test image is a forgery or genuine. Such critical assessments of authenticity are vital in maintaining security and trust in various applications~\cite{wei:19IDN}.

Signature verification is challenging due to several factors. First, people have unique ways of signing. Second, there's often not enough detailed information about how each signature stroke is made. Finally, sophisticated forgeries can be hard to distinguish from genuine signatures. The essence of signature verification lies in comparing the similarity of the subtle stylistic characteristics concealed within the reference signature and the testing one rather than focusing on the specific contents of the signatures~\cite{wei:19IDN}.

Traditional approaches for OSV heavily rely on manual handcrafted feature engineering~\cite{kumar:12, Dutta:16}. In recent years, numerous deep-learning methods have been proposed~\cite{wei:19IDN}. These deep-learning methods have demonstrated significant advancements in verification performance when compared to traditional handcrafted features. However, some key issues still need to be well addressed by previous deep-learning methods. Despite the importance of the reasoning process that considers the similarity between global features from the holistic signature image. They lack the incorporation of structural comparison among local patch features to measure similarity. A global representation destroys image structures and leads to the loss of local information. Local features (stroke structure and style) offer discriminative and transferable information for offline signature verification (\cref{fig_intro}). Hence, a desirable metric-based OSV should possess the capability to leverage local discriminative representations for metric learning while minimizing the influence originating from irrelevant regions.

To address the issues mentioned above, we propose a new model \textbf{DetailSemNet} for offline signature verification. In this model, \textbf{Structural Matching} is proposed to align local patch tokens. This mechanism enhances the model's ability to capture local discriminative features, thereby significantly improving its identification capabilities. While integrating \textbf{Structural Matching} directly into DetailSemNet has been observed to strengthen performance, we noted a crucial limitation in attention/transformer-based models, where they often lose detailed information during the token feature extraction process. Our preliminary analysis, illustrated in \cref{fig_intro_fft}, supports this observation. Traditional transformer-based models primarily focus on low-frequency patterns, neglecting the high-frequency information crucial for distinguishing between similar signatures. Consequently, when tested with images rich in high-frequency details, the Equal Error Rate (EER) performance shows negligible improvement. This observation suggests a performance improvement gap that deserves further exploration.


To this end, as shown in \cref{fig_compare}, we deliberately designed multi-branch networks to extract the Detail and Semantics components and handle them separately during the feature extraction process. This approach allows us to retain more detailed information, resulting in the model exhibiting improved discriminative capabilities. Compared with the conventional transformer-based method, as shown in~\cref{fig_intro_fft}, our approach can well use high-frequency information to enhance system performance. Below, we summarize our contributions.

\begin{enumerate}
    \item Our method introduces \textbf{Structural Matching}, a novel technique designed to optimize the matching of local embeddings. Combined with global distance measures, it enables a more comprehensive assessment of similarity in offline signature verification, enhancing the model's accuracy and reliability.
    \item We propose the \textbf{Detail-Semantics Integrator}, an innovative network that ensures the detail and Semantics features are extracted. This integrator significantly improves the model's feature extraction capabilities and achieves a finer understanding of the signature data, making it particularly well-suited for the OSV task.
    \item Our proposed method exhibits superior performance compared to existing methods. This superiority is evident both in single-dataset testing scenarios and cross-dataset evaluations. Such performance highlights the model's generalization ability.
\end{enumerate}
}

\begin{figure}[t]
  \centering
  \begin{subfigure}{0.81\linewidth}
    \includegraphics[width=\linewidth]{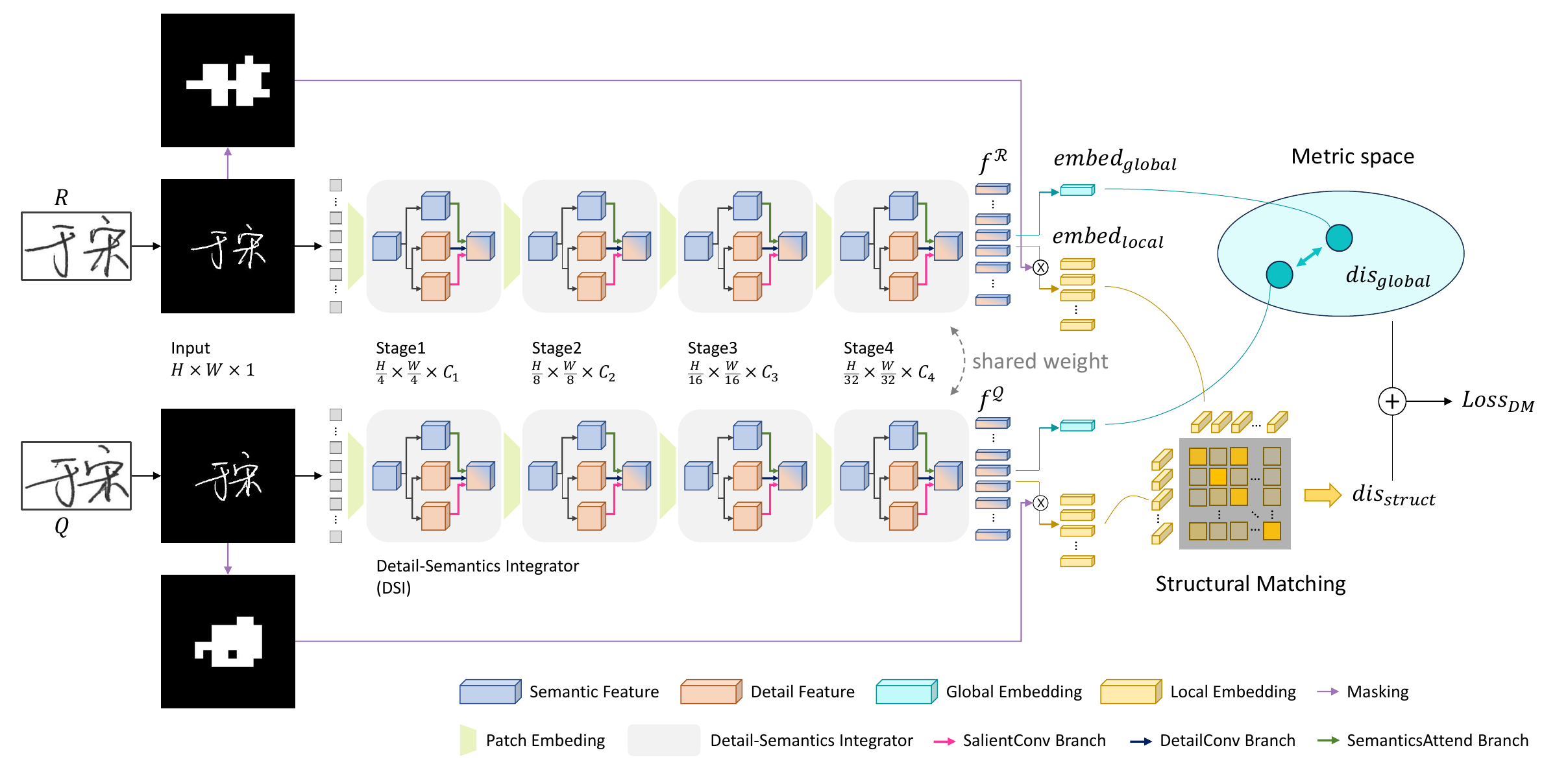}
    \caption{Our DetailSemNet.}
    \label{fig_overview}
  \end{subfigure}
  \hfill
  \begin{subfigure}{0.18\linewidth}
    \includegraphics[width=\linewidth]{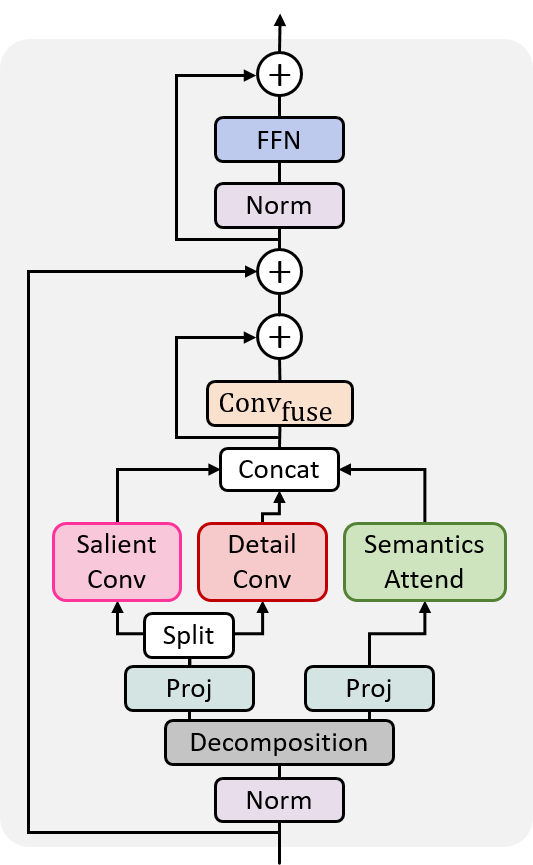}
    \caption{The proposed Detail-Semantics Integrator (DSI).}
    \label{fig_detail}
  \end{subfigure}
  \caption{In our \textbf{DetailSemNet}, the \textbf{Detail-Semantics Integrator} splits  features into two components: Semantic and Detail. The Semantic component is sent to the SemanticsAttend Branch to gather context information, while the Detail component goes through the SalientConv Branch and DetailConv Branch. The outputs of these three branches are then fused, creating features that incorporate both detailed and semantic information. In addition to global similarity, the model also performs structural matching on the features, allowing it to pay more attention to detailed information.}
  \label{fig_}
\end{figure}
\section{Related works}
\label{sec:related}

\subsection{Offline Signature Verification (OSV)}
Offline signature verification has been a subject of study for many years, as evidenced by Dey et al.~\cite{dey2017signet}. More recently, a shift has been observed where deep-learning methods surpass manually-designed feature methods in performance~\cite{kumar:12, Dutta:16}. For instance, Wei et al.\cite{wei:19IDN} improved feature extraction for verification using gray-inverted images and attention modules. Li et al.~\cite{Li:21SDINet} innovatively integrated sequential representations into static signature images, creating a unified framework for offline verification. Further, Li et al.~\cite{Li:22AVN} leveraged an adversarial network to enhance signature verification capabilities. The use of a dual-channel Network to measure image dissimilarity was introduced by Li et al.~\cite{Li:19DeepHSV}. Lately, Lu et al.~\cite{Lu:21Cac} proposed a network which assesses pairs of images using a cycling method to make observations. Additionally, Li et al.~\cite{Li02:22TranOSV, TransOSV_PR} marked a significant milestone by being the first to introduce a transformer framework in this domain.

Distinguishing our work from these prior efforts, we present a novel approach to offline signature verification by directly utilizing local patch tokens to learn a distance metric and explicitly find the patch matching between an image pair, thereby harnessing their potential to enhance the verification process.

\subsection{Local Matching}
\MC{
Developing an effective Local Matching method for OSV is challenging. The Hungarian algorithm~\cite{Hungarian} efficiently solves the Assignment Problem but may not suit signature verification due to its strict one-to-one assignment constraint. Signatures often vary and do not align perfectly. Various distance metrics measure similarity between sets: the Hausdorff distance (HD) \cite{Hausdorff} is sensitive to outliers, the Chamfer distance (CD) \cite{wu2021densityaware} sums squared distances between nearest neighbors, and the Earth Mover's distance (EMD) \cite{wu2021densityaware, zhang:22} finds the least costly way to align sets. While EMD is more detail-sensitive than CD, it is also computationally complex \cite{liu2019morphing}, requiring a balance between efficiency and capturing the nuances of signature variations.
}

\subsection{Backbone Designs and Their Properties}
The field of visual data processing has evolved from Convolutional Neural Networks (CNNs) to Vision Transformer models (ViTs). However, recent studies, like Raghu et al.~\cite{raghu2021do} and Bai et al.~\cite{Bai22}, have explored the strengths and limitations of ViTs compared to CNNs. A key finding is that while ViTs are adept at global information aggregation through self-attention, they might not leverage local image structures as effectively as CNNs and reduce the ability to model local fine-grained details~\cite{Luo_2021_CVPR, Wang2020HFC, Pinto2022AnIT, Li2022LocalityGF}.

To address this, new ViT architectures have been introduced. PVT~\cite{PVT} adopts a pyramid-like structure for diverse resolution feature maps, enhancing its utility in dense prediction tasks. Swin~\cite{Swin} focuses on local window self-attention. DAT~\cite{DAT} incorporates a deformable attention mechanism, offering flexible adaptability for local feature representation. BiFormer~\cite{biformer} uses a Bi-Level Routing Attention mechanism, focusing on semantically rich regions for fine-grained attention.
These advancements underscore the importance of detailed feature extraction and semantic learning, which are vital for Offline Signature Verification (OSV) tasks. Inspired by these insights, we proposed \textbf{DetailSemNet} to combine the advantages of CNNs and ViTs, making it particularly effective for OSV.

\section{Method}
\label{sec:method}


\CCH{
Our model, as illustrated in \cref{fig_overview}, processes a pair of input images, \(R\) (Reference) and \(Q\) (Query), for signature verification. These images undergo preprocessing to convert them into binary images of dimensions \(H \times W \times 1\) coupled with a foreground map, which is then used as inputs for the model. The images are segmented into patches (i.e., tokenized) and fed into the feature extraction backbone, which consists of four stages. Each stage involves a Patch Embedding layer followed by several layers of the proposed \textbf{Detail-Semantics Integrator} (DSI) module (see \cref{subsec:DSI} and \cref{fig_detail}). The backbone's output comprises two sets of token features, \(f^{\mathcal{R}}\) and \(f^{\mathcal{Q}}\), which are utilized to compute two types of distances, \(dis_{global}\) and \(dis_{struct}\). First, as shown in \cref{eqn:global_dist}, we calculate the global \(L_2\) distance \(dis_{global}\) using the global embeddings \(embed^{\mathcal{R}}_{global}\) and \(embed^{\mathcal{Q}}_{global}\), derived by averaging the feature set \(f^{\mathcal{R}}\) and set \(f^{\mathcal{Q}}\) accordingly. Second, our Structural Matching technique is applied to ascertain a local structural distance \(dis_{struct}\), computed from the local embeddings \(embed^{\mathcal{R}}_{local}\) and \(embed^{\mathcal{Q}}_{local}\) as outlined in \cref{eqn:struct_dist}. The process for calculating \(dis_{struct}\) is detailed in \cref{subsec:SM}.

\begin{equation}
dis_{global} = L_2(embed^{\mathcal{R}}_{global}, embed^{\mathcal{Q}}_{global})
\label{eqn:global_dist}
\end{equation}
In the evaluation phase, we utilize the combined distance, defined in \cref{eqn:dist}, as the similarity measurement of \(R\) and \(Q\) and make a final verification decision by thresholding. 
\begin{equation}
dis = \lambda_0 \times dis_{global} + dis_{struct},
\label{eqn:dist}
\end{equation}
where \(\lambda_0\) is a hyperparameter, used to adjust the weighting between the two distances.

\subsection{Local Structural Matching}
\label{subsec:SM}
To implement Structural Matching, we focus on utilizing features from image regions containing drawn strokes for local similarity measurement and filtering out the background tokens according to the foreground map. This step is particularly beneficial due to the sparsity of signature images, where some local patches are uninformative. In addition, we mask out the background tokens after our feature extraction backbone to reduce the information loss. 

To extract the foreground map, we first resize the input image from \(H \times W \times 1\) to  \(h \times w\) to match the size of the extracted token feature sets (i.e., \(f^{\mathcal{R}}\) and \(f^{\mathcal{Q}}\)). Following this, we apply global thresholding to binarize the resized image and achieve the foreground map, \(Mask\), which delineates the relevant regions containing strokes. It's important to note that a more sophisticated text region segmentation technique can be employed when the input images have cluttered backgrounds. 
Later, this \(Mask\) is applied to filter out irrelevant tokens from the entire token set, leaving us with significant tokens. These tokens are subsequently processed through a linear layer to generate the local embeddings \(embed_{local}\), which are crucial for calculating the local structural distance \(dis_{struct}\). Specifically, we define the corresponding masks of \(R\) and \(Q\) as \(Mask^{\mathcal{R}}\) and \(Mask^{\mathcal{Q}}\). The local embeddings of \(R\) and \(Q\) and their structural distance \(dis_{struct}\) can be calculated by  
\begin{equation}
embed^{\mathcal{R}}_{local} = linear(Mask^{\mathcal{R}}(f^{\mathcal{R}})),
\end{equation}
\begin{equation}
embed^{\mathcal{Q}}_{local} = linear(Mask^{\mathcal{Q}}(f^{\mathcal{Q}})),~and 
\end{equation}
\begin{equation}
dis_{struct} = SM(embed^{\mathcal{R}}_{local}, embed^{\mathcal{Q}}_{local}).
\label{eqn:struct_dist}
\end{equation}



In \cref{eqn:struct_dist}, to effectively match local structures, or tokens, we developed the function \({SM}\) to quantify the similarity between two sets of local embeddings, \(embed^{\mathcal{R}}_{local} = \{r_0, r_1, ..., r_{N-1}\}\) and \(embed^{\mathcal{Q}}_{local} = \{q_0, q_1, ..., q_{M-1}\}\), where $N$ and $M$ represent the number of embeddings remaining after masking. To determine the local distance \(dis_{struct}\), we initially calculate the cosine distance between pairs of local embeddings by
\begin{equation}
d_{ij} = 1 - \frac{r_i^Tq_j}{\|r_i\|\|q_j\|},~ 0\leq i\leq N-1~and~0\leq j\leq M-1.
\label{eqn:ground_dist}
\end{equation}
This calculation forms the basis of the ground distance matrix \(D = (d_{ij}) \in \mathbb{R}^{N \times M}\), which is used for token matching.



The matching relationship between the tokens within the two local embedding sets is essential to obtain \(dis_{struct}\). To represent the matching relationship, we introduce a flow matrix \(\emph{F} = [f_{ij}] \in \mathbb{R}^{N \times M}\), where each element \(f_{ij}\) indicates the ratio of the matching assignment from \(r_i\) to \(q_j\). Note that \(r_i\) can match to multiple \(q_j\). Once the optimal flow matrix \(\emph{F}^* = [f^*_{ij}] \) (i.e., representing the ideal matching relationships) is determined, we can then compute a more accurate similarity measure. As defined in \cref{eqn:emd}, this is achieved by performing an element-wise multiplication of the ground distance matrix \(D = (d_{ij})\) with the optimal flow matrix \(\emph{F}^* = [f^*_{ij}] \), followed by summing up these products and normalizing the result. The outcome of this process is a refined similarity measure that more accurately reflects the proper correspondence between the two sets of local tokens.
\begin{equation}
dis_{struct} = \frac{\sum_{i=0}^{N-1}\sum_{j=0}^{M-1}d_{ij}f^*_{ij}}{\sum_{i=0}^{N-1}\sum_{j=0}^{M-1}f^*_{ij}}.
\label{eqn:emd}
\end{equation}

Here, we conceptualize the problem of determining the optimal matching flow, \(\emph{F}^* = [f^*_{ij}] \), as the process of minimizing the Earth Mover's Distance (EMD) \cite{Rubner:00}, whose details would be provided in the supplementary. The EMD framework allows us to assign different weights to each local embedding, given the total weights of all tokens sum up to 1. In this context, the weight of each local embedding represents its significance in the matching process. Our experiments find that a simple uniform weight can also achieve good results. However, if we have the prior information, we can adjust the weights for further improvement. Next, to discover the optimal matching flow, we reformulate the EMD optimization problem as a linear programming challenge and utilize the Sinkhorn algorithm \cite{sinkhorn} for problem-solving. The Sinkhorn algorithm smoothens the EMD calculation through entropic regularization, making it possible to solve this linear programming problem effectively. Once the optimal matching flow, \(\emph{F}^* = [f^*_{ij}] \), has been calculated, we can calculate \(dis_{struct}\) by \cref{eqn:emd}. 


\subsection{Detail-Semantics Integrator}
\label{subsec:DSI}

Although transformer-based models have been widely adopted, several studies have discussed how the Multi-head Self-Attention module (MSA) indiscriminately suppresses high-frequency signals, leading to significant information loss. These discussions include approaches such as \cite{wang2022antioversmoothing, Bai22, park2022how, Zheng2023PreservingLI}. This is not advantageous for tasks like OSV. Addressing this, as depicted in \cref{fig_detail}, we have developed the \textbf{Detail-Semantics Integrator
}(DSI), a novel feature enhancement technique tailored for feature maps. The DSI begins by decomposing the input feature \(X\) into two parts: the Semantic Feature \(Sem[X]\) and Detailed Feature \(Det[X]\). To maintain computational efficiency, we compute \(Sem[X]\) by performing local average pooling on \(X\), which effectively captures lower-frequency parts. Conversely, \(Det[X]\), representing higher-frequency parts, is computed by \(X - Sem[X]\). Later, after a 1-by-1 projection layer, \(Sem[X]_{proj}\) undergoes processing in the SemanticsAttend Branch \(SemAtt\), using an attention-based module to extract semantic features \(Y_{Sem}\). On the other hand, \(Det[X]_{proj}\) is processed through a convolution-based module \(Conv\) to extract fine-grain details \(Y_{Det}\) from local features. The steps are as follows:
\begin{equation}
    Y_{Sem} = SemAtt(Sem[X]_{proj}) ~and
\end{equation}
\begin{equation}
    Y_{Det} = Conv(Det[X]_{proj}).
\end{equation}
Moreover, the convolution-based module is strategically divided into the \textit{SalientConv} branch and the \textit{DetailConv Branch}. The \textit{SalientConv} branch integrates both maximum filter and convolution layers, where the maximum filter is particularly effective in retaining salient features and helps to highlight the prominent aspects of the signature images.
Conversely, the \textit{DetailConv} branch, consisting of two successive convolution layers, is tailored to draw out finer details. Unlike attention mechanisms, convolutions excel at detecting intricate high-frequency details, making them ideal for processing the \(Det[.]\) part of the input. For feature decomposition, the projected detailed features \(Det[X]_{proj}\) are divided equally along the channel direction, with each part directed to either \(SalientConv\) or \(DetailConv\) for specialized processing. The detailed steps are as follows:
\begin{equation}
    Det1[X], Det2[X] = split(Det[X]_{proj}).
\end{equation}
\begin{equation}
    Y_{Det1} = SalientConv(Det1[X]).
\end{equation}
\begin{equation}
    Y_{Det2} = DetailConv(Det2[X]).
\end{equation}
In the final stage of DSI (\cref{fig_detail}), the outputs from the \(SemAtt\), \(SalientConv\), and \(DetailConv\) branches are concatenated along the channel dimension to form a unified feature representation, followed by a Residual Convolution layer (\(Conv_{fuse}\)). This layer further integrates the concatenated features to ensure a seamless blend of detailed and semantic information. Moreover, consistent with the architecture of most transformer-based models, DSI incorporates a Feed-Forward Network (FFN) and Layer Normalization (LN) to enhance the model's processing capabilities. In \cref{subsec:backbone}, we demonstrate through experiments that this particular design of DetailSemNet is highly effective for OSV tasks.


\subsection{Loss Function}

The output of our signature verification model is \emph{dis} (\cref{eqn:dist}), which denotes the distance between two input signature images. To train the model, we employ a double-margin contrastive loss~\cite{Lu:21Cac, Hao2018DeepFirearmLD}, defined as follows: 
\begin{equation}
Loss_{DM} = y\{max(0, dis - m)\}^2 + (1 - y)\{max(0, n - dis)\}^2.
\end{equation}
Here, the supervised label \(y\) is assigned a value of 1 for positive (genuine-genuine) signature pairs and 0 for negative (genuine-forged) signature pairs. The parameters \(n\) and \(m\) represent the margin values used in the loss calculation. Importantly, \(m\) is constrained to be less than \(n\) to ensure the loss function behaves as intended. 
}


\section{Experiments}
\label{sec:expe}
\CCH{
\begin{table}[t]
  \caption{Signature verification comparison on BHSig-H(\%) and BHSig-B(\%).}
  \centering
  \begin{tabular}{l||rrrr|rrrr}
    \toprule
    \multirow{2}{*}{Method} & \multicolumn{4}{c|}{BHSig-H} & \multicolumn{4}{c}{BHSig-B} \\
     & ~FAR & ~FRR & ~Acc $\uparrow$ & ~EER $\downarrow$ & ~FAR & ~FRR & ~Acc $\uparrow$ & ~EER $\downarrow$ \\
    \midrule
    SigNet\cite{dey2017signet}    & 15.36 & 15.36 & 84.64 & 15.36 & 13.89 & 13.89 & 86.11 & 13.89 \\
    IDN\cite{wei:19IDN}           &  4.93 &  8.99 & 93.04 &  6.96 &  4.12 &  5.24 & 95.32 &  4.68 \\
    DeepHsv\cite{Li:19DeepHSV}    &     - &     - & 86.66 & 13.34 &     - &     - & 88.08 & 11.92 \\
    SDINet\cite{Li:21SDINet}      &  6.24 &  3.77 & 95.00 &  5.11 &  3.30 &  7.86 & 94.42 & 5.39 \\
    CaC\cite{Lu:21Cac}            &  5.97 &  5.97 & 94.03 &  5.97 &  3.96 &  3.96 & 96.04 & 3.96 \\
    AVN\cite{Li:22AVN}            &  5.46 &  5.91 & 94.32 &  5.65 &  7.33 &  5.07 & 93.80 & 6.14 \\
    TransOSV\cite{Li02:22TranOSV, TransOSV_PR} &  3.39 &  \textbf{3.39} & 96.61 &  3.39 &  9.95 &  9.95 & 90.05 & 9.95 \\
    2C2S\cite{REN2023105639}      &  8.66 &  5.16 & 90.68 &  9.32 &  5.37 &  8.11 & 93.25 & 6.75  \\
    SURDS\cite{SURDS}             & 12.01 &  8.98 & 89.50 &     - & 19.89 &  5.42 & 87.34 &    -  \\
    MA-SCN\cite{Zhang2022MASCN}   &  5.73 &  4.86 & 94.99 &  5.32 &  9.96 &  5.85 & 92.86 & 8.18 \\
    SigGCN\cite{Ren2023VisionGC}  & 12.96 & 11.27 & 87.88 & 12.17 &  4.06 &  \textbf{3.95} & 95.99 & 4.00 \\
 Co-Tuplet\cite{Huang2023MultiscaleFL} & 6.76 & 6.56 & - & 6.68 & 5.93 & 6.20 & - & 6.12 \\
 HybridFE\cite{Xiong2023HybridFE} & 11.74 & 11.74 & 88.26 & 11.74 & 8.36  & 8.36  & 91.64 & 8.36 \\
    SPD Manifold\cite{Zois}       &     - &     - &     - & 15.60 &     - &     - &     - & 11.10 \\
    \textbf{Ours}                 &  \textbf{1.07} &  3.59 & \textbf{98.24} &  \textbf{2.07} &  \textbf{0.95} &  4.04 & \textbf{98.19} & \textbf{2.11} \\
    \bottomrule
  \end{tabular}
  \label{tab:table_BHSig_B_H}
\end{table}
We evaluated our approach using four challenging datasets, each representing a different language. These include the CEDAR Dataset~\cite{CEDAR} (English), BHSig-B Dataset~\cite{BHSig} (Bengali), BHSig-H Dataset~\cite{BHSig} (Hindi), and ChiSig Dataset\footnote{The latest ChiSig Dataset, unlike the others, has not been previously utilized for Offline Signature Verification (OSV) tasks. Therefore, it was specifically used in our ablation study to underscore the efficacy of our method.}~\cite{Yan_2022_CVPR} (Chinese). To further test the robustness of our model, we conducted cross-language experiments. These involved training the model on a dataset and subsequently testing it on a dataset in a different language. Our model was trained on an NVIDIA GeForce RTX 2080Ti GPU. The initial model pre-trained weights were derived from the ImageNet1K dataset~\cite{deng2009imagenet}.

We performed comprehensive comparisons with several existing methods \cite{Li:21SDINet, wei:19IDN, Li02:22TranOSV} using a variety of metrics. These metrics include the False Rejection Rate (FRR), False Acceptance Rate (FAR), Equal Error Rate (EER), Area Under the Curve (AUC), and Accuracy (Acc). Furthermore, we utilize the Equal Error Rate (EER) to identify the point where FRR and FAR are equal. This equilibrium point informs the threshold used to calculate Accuracy (Acc) and facilitates the verification decision-making process.
\subsection{Results on BHSig-B and BHSig-H Datasets}

The BHSig260 Dataset\cite{BHSig} includes both the BHSig-B and BHSig-H Datasets. The BHSig-B Dataset contains signatures from 100 individuals from Bengal, with each individual contributing 24 genuine signatures and 30 forgeries. For our model's training, we utilized the signatures from 50 of these individuals, while the signatures from the remaining 50 were reserved for testing purposes. In contrast, the BHSig-H Dataset consists of signatures from 160 individuals, with each providing 24 genuine signatures and 30 forged signatures. In this case, our training involved signatures from 100 individuals, with the remaining 60 individuals' signatures set aside for the testing phase.

Our mode performance was compared with several conventional approaches, including SigNet \cite{dey2017signet}, IDN \cite{wei:19IDN}, DeepHsv \cite{Li:19DeepHSV}, SDINet \cite{Li:21SDINet}, CaC \cite{Lu:21Cac}, AVN \cite{Li:22AVN}, TransOSV \cite{Li02:22TranOSV, TransOSV_PR}, 2C2S \cite{REN2023105639}, SURDS \cite{SURDS}, MA-SSN \cite{Zhang2022MASCN}, SigGCN \cite{Ren2023VisionGC}, HybridFE\cite{Xiong2023HybridFE}, and SPD Manifold \cite{Zois}. The evaluation results are detailed in \cref{tab:table_BHSig_B_H}.

Regarding the BHSig-B Dataset, our method demonstrated impressive results, achieving an Accuracy (Acc) of 98.19\%, a False Acceptance Rate (FAR) of 0.95\%, and a False Rejection Rate (FRR) of 4.04\%. Additionally, the model recorded an Equal Error Rate (EER) of 2.11\%. Compared to the best available result from other methods, our approach exhibits a significant performance gain of 1.85\%. Turning to the BHSig-H Dataset, our method continued its strong performance, achieving an Accuracy of 98.24\%, an FAR of 1.07\%, and an FRR of 3.59\%. The EER for this dataset was an impressive 2.07\%. Compared with the leading comparative methods, our approach marks a notable performance improvement of 1.32\%.

\setlength\intextsep{0pt}
\begin{wraptable}[16]{r}{0.6\linewidth}
\caption{OSV comparison on CEDAR(\%).}
  \centering
  \begin{tabular}{l||rrrr}
    \toprule
    Method & ~FAR & ~FRR & ~Acc $\uparrow$ & ~EER $\downarrow$ \\
    \midrule
    SigNet\cite{dey2017signet} & -    & -    & -     & 4.63 \\
    IDN\cite{wei:19IDN}        & 5.87 & 2.17 & -     & 3.62 \\
    SDINet\cite{Li:21SDINet}   & 3.42 & 0.73 & -     & 1.75 \\
    CaC\cite{Lu:21Cac}         & 4.34 & 4.34 & 95.66 & 4.43 \\
    AVN\cite{Li:22AVN}         & 3.26 & 4.42 & 96.16 & 3.77 \\
 MA-SCN\cite{Zhang2022MASCN}   & 19.21 & 18.35 & 80.75 & 18.92 \\
    MLDF\cite{Arab}            &    - &    - &     - & 5.00 \\
 Co-Tuplet\cite{Huang2023MultiscaleFL} & 3.33 & 3.55 & - & 3.51 \\
 HybridFE\cite{Xiong2023HybridFE} & 9.95 & 9.95 & 90.05 & 9.95 \\
    SPD Manifold\cite{Zois}    &    - &    - &     - & 8.53 \\
    \textbf{Ours}              & \textbf{0.36} & \textbf{0.58} & \textbf{99.53} & \textbf{0.58} \\
    \bottomrule
  \end{tabular}
  \label{tab:table_CEDAR}
\end{wraptable}
\subsection{Results on CEDAR Dataset}
Within the CEDAR signature dataset~\cite{CEDAR}, each individual is represented by 24 genuine and 24 forged signatures, all written in English. Aligning with methodologies from previous studies, we used the signatures of 50 individuals to train our model and reserved the signatures of the remaining 5 individuals for testing. In this setup, a positive sample is created by pairing a reference signature with a genuine signature, while a negative sample is formed by pairing a reference signature with a forged signature. Thus, each signatory contributes 276 positive and 276 negative pairs for verification.

We conducted a comparative analysis of our model with other conventional approaches such as SigNet~\cite{dey2017signet}, IDN~\cite{wei:19IDN}, SDINet~\cite{Li:21SDINet}, CaC~\cite{Lu:21Cac}, AVN~\cite{Li:22AVN}, MA-SCN \cite{Zhang2022MASCN}, MLFD \cite{Arab}, Co-Tuplet \cite{Huang2023MultiscaleFL}, HybridFE\cite{Xiong2023HybridFE}, and SPD Manifold \cite{Zois}. The results of these comparative analyses on the CEDAR Dataset are detailed in~\cref{tab:table_CEDAR}. Here, our method demonstrated exemplary performance, achieving an accuracy of 99.53\%, a False Acceptance Rate (FAR) of 0.36\%, and a False Rejection Rate (FRR) of 0.58\%. Additionally, the model attained an Equal Error Rate (EER) of 0.58\%. Compared to the best results from other methods, our approach appeared as a top-performing model.




\subsection{Verification on Cross-Dataset Scenario}

To evaluate the generalization capabilities of our model, 
we train the model on a dataset and directly test the model on other datasets with different languages.
The effectiveness of our model in this regard is demonstrated by the test results presented in \cref{tab:table_Cross}. Compared to other methods, our approach consistently outperforms them, indicating its adaptability and robustness across different languages without model finetune. 

\begin{table}[t]
\caption{The zero-shot cross-lingual OSV task (cross-dataset) testing results. (EER\%)} 
\centering
\begin{tabular}{c||rr|rr|rr}
\toprule
Train & \multicolumn{2}{c|}{BHSig-H} & \multicolumn{2}{c|}{BHSig-B} & \multicolumn{2}{c}{CEDAR} \\
\midrule
Test & \multicolumn{1}{c}{~BHSig-B} & \multicolumn{1}{c|}{~CEDAR} & \multicolumn{1}{c}{~BHSig-H} & \multicolumn{1}{c|}{~CEDAR} & \multicolumn{1}{c}{~BHSig-H} & \multicolumn{1}{c}{~BHSig-B} \\ 
\midrule
\multicolumn{1}{l||}{SigNet~\cite{dey2017signet}} & 39.35 & 40.43 & 35.43 & 50.00 & 44.39 & 35.85 \\
\multicolumn{1}{l||}{IDN~\cite{wei:19IDN}} & 25.88 & 50.00 & 25.70 & 50.00 & 49.64 & 49.99 \\
\multicolumn{1}{l||}{CaC~\cite{Lu:21Cac}} & 14.66 & 29.49 & 30.41 & 33.71 & 39.08 & 38.07 \\
\multicolumn{1}{l||}{SURDS~\cite{SURDS}} & 27.74 & - & 32.99 & - & - & - \\
\multicolumn{1}{l||}{TransOSV~\cite{Li02:22TranOSV, TransOSV_PR}} & 18.66 & - & 17.17 & - & - & - \\
\multicolumn{1}{l||}{\textbf{Ours}} & \textbf{7.46} & \textbf{14.05} & \textbf{15.91} & \textbf{7.32} & \textbf{16.35} & \textbf{8.40} \\
\bottomrule
\end{tabular}
\label{tab:table_Cross}
\end{table}
\begin{table*}
\caption{Ablation study of Structural Matching (SM), DetailConv Branch (DCB), and SalientConv Branch (SCB). The table presents the results using the abbreviations M, D, and S in that order. Four different datasets are tested.}
\centering
    \begin{tabular}{ccc||rrr|rrr|rrr|rrr}
        \toprule
        \multirow{2}{*}{M} & \multirow{2}{*}{D} & \multirow{2}{*}{S} & \multicolumn{3}{c}{BHSig-H} & \multicolumn{3}{|c}{BHSig-B} & \multicolumn{3}{|c}{CEDAR} & \multicolumn{3}{|c}{ChiSig} \\
        & & & EER & AUC & Acc & EER & AUC & Acc & EER & AUC & Acc & EER & AUC & Acc \\
        \midrule
        $\times$ & $\times$ & $\times$ & 4.70 & 0.991 & 95.82 & 3.37 & 0.995 & 97.14 & 3.41 & 0.994 & 96.81 & 12.47 & 0.947 & 88.68 \\
        \checkmark & $\times$ & $\times$ & 4.67 & 0.992 & 95.88 & 3.29 & 0.995 & 97.22 & 1.99 & 0.998 & 98.08 & 10.69 & 0.964 & 89.82 \\
        $\times$ & \checkmark & $\times$ & 2.72 & 0.997 & 97.70 & 2.51 & 0.997 & 97.44 & 1.45 & 0.999 & 98.77 & 8.91 & 0.972 & 91.73 \\
        $\times$ & $\times$ & \checkmark & 2.87 & 0.997 & 97.69 & 2.66 & 0.997 & 97.68 & 1.59 & 0.998 & 98.41 & 8.65 & 0.977 & 92.62 \\
        $\times$ & \checkmark & \checkmark & 2.62 & 0.997 & 97.80 & 2.50 & 0.997 & 97.89 & 1.74 & 0.999 & 98.59 & 7.00 & \textbf{0.985} & 93.89 \\
        \checkmark & \checkmark & $\times$ & 2.51 & 0.997 & 97.87 & 2.19 & \textbf{0.998} & 98.03 & 1.09 & 0.999 & 98.95 & 8.65 & 0.977 & 91.35 \\
        \checkmark & $\times$ & \checkmark & 2.72 & 0.997 & 97.74 & 2.19 & \textbf{0.998} & 98.15 & 2.10 & 0.998 & 98.19 & 6.36 & 0.983 & 93.89 \\
        \checkmark & \checkmark & \checkmark & \textbf{2.07} & \textbf{0.998} & \textbf{98.24} & \textbf{2.11} & \textbf{0.998} & \textbf{98.19} & \textbf{0.58} & \textbf{1.000} & \textbf{99.53} & \textbf{5.85} & \textbf{0.985} & \textbf{94.40} \\
        \bottomrule
    \end{tabular}
    \label{tab:table_ablation_dataset}
\end{table*}
\subsection{Ablation Studies}

We conducted ablation experiments to evaluate the impact of each module introduced in our proposed method, including Structural Matching (SM), DetailConv Branch (DCB) and SalientConv Branch (SCB). These tests were performed across four different datasets, and the results are presented in \cref{tab:table_ablation_dataset}. Through these experiments, we observed a consistent trend of progressive performance improvement with the sequential incorporation of each proposed modification. 

To further understand the impact of our proposed Structural Matching, we experimented with its integration at various stages within the model. The outcomes of these tests are detailed in \cref{tab:ablation_diff_stage}. Notably, the results demonstrate that integrating Structural Matching towards the end of the process yields the most favorable performance.}

\begin{table}[ht]
\centering
\begin{minipage}[t]{0.35\linewidth} 
    \centering
    \caption{Apply Structural Matching to different stages on the BHSig-H.}
        \begin{tabular}{l || r}
            \toprule
            stage & EER(\%) $\downarrow$ \\
            \midrule
            3                  & 3.47          \\
            3 \& 4             & 3.05          \\ 
            \textbf{4 (Ours)}  & \textbf{2.09} \\
            \bottomrule
        \end{tabular}
    \label{tab:ablation_diff_stage}
\end{minipage}
\quad
\begin{minipage}[t]{0.6\linewidth} 
    \centering
    \caption{Comparison of Different Backbones. We test our results on the BHSig-H and BHSig-B. All the models are under identical conditions.}
        \begin{tabular}{l || r r | r r}
            \toprule
            \multirow{2}{*}{Dataset} & \multicolumn{2}{c|}{BHSig-H} & \multicolumn{2}{c}{BHSig-B} \\
             & EER & Acc & ERR & Acc \\
            \midrule
            PVT\cite{PVT} & 4.62 & 96.06 & 2.72 & 97.55 \\
            Swin\cite{Swin} & 4.24 & 96.15 & 10.27 & 91.01 \\
            SPACH\cite{SPACH} & 3.71 & 96.67 & 3.30 & 96.90 \\
            DAT\cite{DAT} & 4.94 & 95.73 & 20.09 & 82.77 \\
            BiFormer\cite{biformer} & 4.38 & 96.10 & 8.66 & 92.38 \\
            \textbf{Ours} &\textbf{2.07} & \textbf{98.24} & \textbf{2.11} & \textbf{98.19}\\
            \bottomrule
        \end{tabular}
    \label{tab:table_different_backbone}
\end{minipage}
\end{table}
\subsection{Comparison of Different Backbones }
\label{subsec:backbone}
We also evaluate the performance impact of various transformer backbones on our model. We conducted training under identical conditions using different transformer architectures, including PVT\cite{PVT}, Swin\cite{Swin}, SPACH\cite{SPACH}, DAT\cite{DAT}, and BiFormer\cite{biformer}, for comparison. The results, illustrating how each backbone influenced the model's performance, are presented in \cref{tab:table_different_backbone}. Compared with other backbones, our model exhibited better performance, underscoring the effectiveness of our design.

\setlength\intextsep{0pt}
\begin{wraptable}[9]{r}{0.47\linewidth}
\caption{Comparison with Re-ID models on BHSig-H.}
  \centering
  \begin{tabular}{l||rrrr}
    \toprule
    Method & ~FAR & ~FRR & ~Acc $\uparrow$ & ~EER $\downarrow$ \\
    \midrule
    BPB\cite{ni2023part} & 2.73 & 7.25 & 96.02  & 4.42 \\
    PAT\cite{Somers_2023}        & 4.95 & 13.32 & 92.73  & 7.85 \\
    \textbf{Ours}              & \textbf{1.07} &  \textbf{3.59} & \textbf{98.24} &  \textbf{2.07} \\
    \bottomrule
  \end{tabular}
  \label{tab:table_reid}
\end{wraptable}

\subsection{Comparing with Verification Tasks}

Verification Tasks like OSV, Re-ID, and face verification aim to assess image similarity but also face distinct technique challenges. \Eg, a Re-ID task contends with pose variations\cite{TransferrableReID}, occlusions \cite{Somers_2023}, or non-discriminative appearance issues \cite{ni2023part}. In contrast, face verification challenges stem from resolution differences\cite{CRFR}, aging appearance\cite{Age_face_verification}, and wearing accessories\cite{Phan:22}. Our method highlights the need for OSV tasks to discern fine-grained differences in signature pairs, a challenge that is less prominent in other tasks.
We applied state-of-the-art Re-ID models to the OSV task, and the results are shown in \cref{tab:table_reid}.

\subsection{Visualize Matching Results}

\begin{figure}[t]
\centering
\includegraphics[width=\columnwidth]{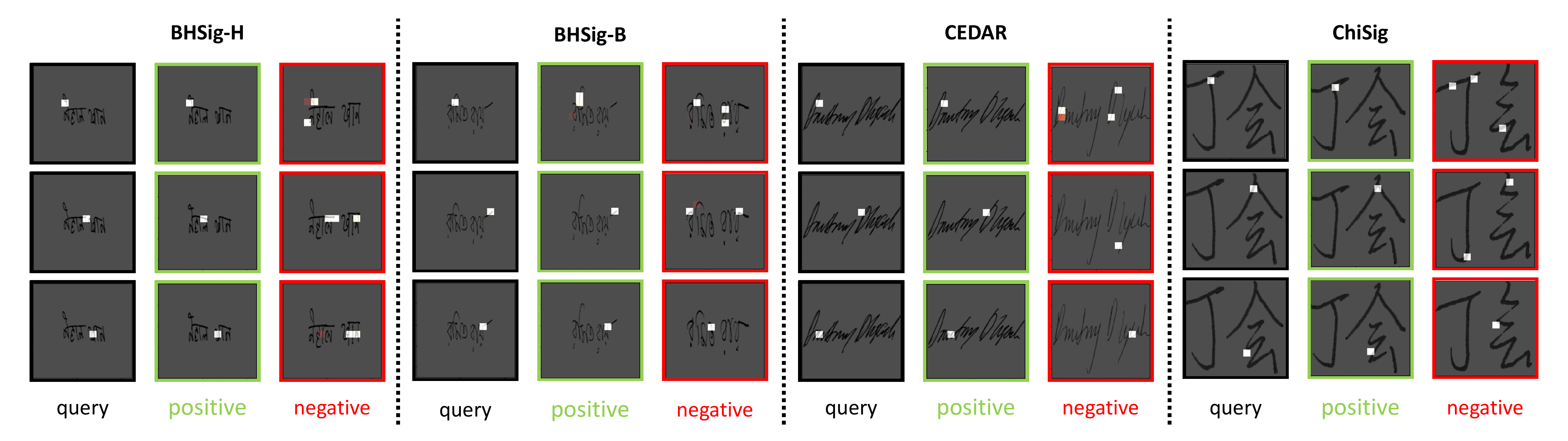}
\caption{Illustrating the matching results of our model on signature pairs. The sample pairs are selected from the four datasets. Our model demonstrates correct matching results when tested on positive pairs; whereas, when tested on negative pairs, it exhibits matching at incorrect positions, sometimes even at multiple locations.}
\label{fig_visualization}
\end{figure}

\begin{figure}
    \centering
    \begin{minipage}{0.48\textwidth}
    \centering
    \includegraphics[width=\linewidth]{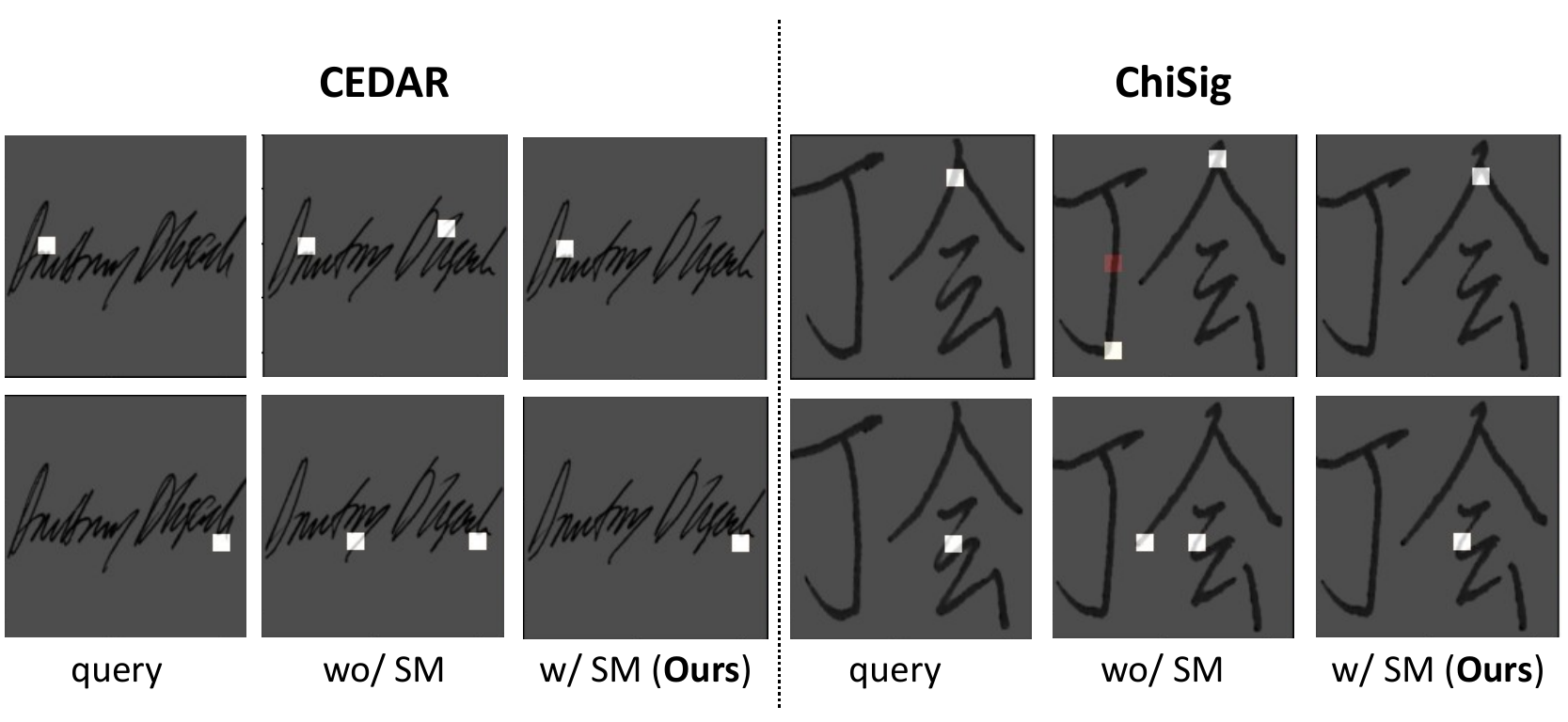} 
    \caption{This figure illustrates the matching results of our model on signature pairs with or without our Structural Matching (SM). The results demonstrates how our Structural Matching improves the model's ability to capture detailed features.}
    \label{fig_visualization_SM}
    \end{minipage}
    \quad
    \begin{minipage}{0.44\textwidth}
    \centering
    \includegraphics[width=\linewidth]{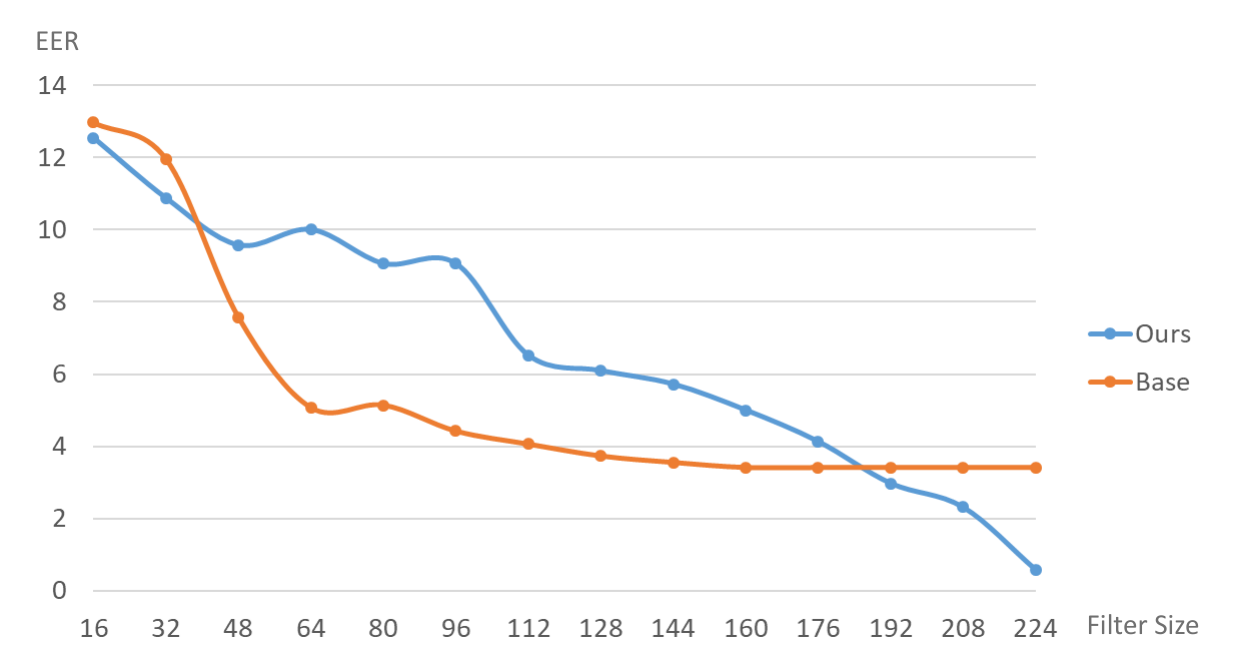}
    \caption{The X-axis shows the amount of high-frequency information in the testing images, with higher values indicating greater inclusion. This graph illustrates that our approach efficiently leverages high-frequency details to reduce EER.}
    \label{fig_filter}
    \end{minipage}
\end{figure}

\MC{
Visualizing matching flows during inference adds interpretability to our model, as shown in \cref{fig_visualization}.
The reference image is placed on the left. It queries a specific patch and then demonstrates how it corresponds to a positive image in the middle and a negative image on the right.Positive pairs typically show correct patch correspondences, while negative pairs struggle with mismatches. We also visualized the impact of Structural Matching in \cref{fig_visualization_SM}, showing that models without it have difficulty achieving precise patch matches.
}



\subsection{Impact of High-Frequency Information on Performance}
\label{subsec:LPF}
\MC{
We use low-pass filters to create testing images with varying high-frequency (HF) content, evaluating both our model and a conventional transformer-based method. The results in \cref{fig_filter} show the impact of HF details on performance, with the X-axis indicating the amount of HF information. Higher HF content generally reduces EERs. For the transformer-based model, EER plateaus at 3.41\% beyond a HF threshold of 164, showing a saturation point. In contrast, our model significantly reduces EER from 6.52\% to 0.58\%, as HF content increases from 112 to 224.
}

\CCH{
\section{Conclusion}
\label{sec:conclud}

In this paper, we introduce \textbf{DetailSemNet}, a novel model for Offline Signature Verification (OSV) that emphasizes local patch features in Structural Matching, a shift from traditional holistic approaches. \textbf{DetailSemNet} also incorporates the Detail-Semantics Integrator (DSI) to enhance structural matching, effectively capturing detailed and semantic aspects. Our results demonstrate that \textbf{DetailSemNet} outperforms existing methods in both single-dataset and cross-dataset scenarios, highlighting its strong generalization capability and potential for real-world application. These findings indicate the effectiveness of combining the DSI module with Structural Matching in OSV models, positioning \textbf{DetailSemNet} as a significant advancement in forensic technology.
}



\section*{Acknowledgements}
This work was supported in part by E.SUN Financial Holding, and financially supported in part (project number: 112UA10019) by the Co-creation Platform of the Industry Academia Innovation School, NYCU, under the framework of the National Key Fields Industry-University Cooperation and Skilled Personnel Training Act, from the Ministry of Education (MOE) and industry partners in Taiwan.  It also supported in part by the National Science and Technology Council, Taiwan, under Grant NSTC-112-2221-E-A49-089-MY3, Grant NSTC-110-2221-E-A49-066-MY3, Grant NSTC-111- 2634-F-A49-010, Grant NSTC-112-2425-H-A49-001-, and in part by the Higher Education Sprout Project of the National Yang Ming Chiao Tung University and the Ministry of Education (MOE), Taiwan. 
(Corresponding author: Ching-Chun Huang.)


%
%
\bibliographystyle{splncs04}
\bibliography{main}
\end{document}


\newcommand{\CCH}[1]{{\color{black}#1}\normalfont} 
\newcommand{\MC}[1]{{\color{black}#1}\normalfont} 
\title{DetailSemNet: Elevating Signature Verification through Detail-Semantic Integration (Supplementary Material)} 

\titlerunning{Elevating Signature Verification through Detail-Semantic Integration}

\author{
Meng-Cheng Shih\inst{1}\orcidlink{0009-0007-4147-6960} \and
Tsai-Ling Huang\inst{1} \and
Yu-Heng Shih\inst{1} \and
Hong-Han Shuai\inst{1}\orcidlink{0000-0003-2216-077X} \and
Hsuan-Tung Liu\inst{2} \and
Yi-Ren Yeh\inst{3} \and
Ching-Chun Huang\inst{1}\orcidlink{0000-0002-4382-5083}
}

\authorrunning{Shih et al.}

\institute{National Yang Ming Chiao Tung University, Taiwan \email{\{mcshih.ee11,christina.ii12,ra890927.cs12,hhshuai,chingchun\}@nycu.edu.tw} \and
E.SUN Financial Holding Co., Ltd, Taiwan \email{ahare-18342@esunbank.com.tw}\and
National Kaohsiung Normal University, Taiwan
\email{yryeh@nknu.edu.tw}}

\maketitle
\setcounter{page}{1}

\setcounter{section}{0}
\setcounter{table}{0}
\renewcommand{\thetable}{A\arabic{table}}
\setcounter{equation}{0}
\renewcommand{\theequation}{A\arabic{equation}}
\setcounter{figure}{0}
\renewcommand{\thefigure}{A\arabic{figure}}


\MC{
\section{More Comparison Results}
In addition to the related works listed in Tab. 1 and Tab. 2, we extend our comparison by including several state-of-the-art (SOTA) methods, namely SET~\cite{Ren2022SETAS}, SWIS~\cite{SWIS}, SigCNN~\cite{Jiang2022ForgeryfreeSV}, and Consensus~\cite{Brimoh2024ConsensusThresholdCF}, using the BHSig-H, BHSig-B, and CEDAR Datasets. These additional comparisons are presented in \cref{tab:table_BHSig_B_H_more} and \cref{tab:table_CEDAR_more}, enhancing the comprehensiveness of our performance evaluation. The results further demonstrate that our method surpasses previous approaches.

\begin{table}
  \caption{More OSV comparison on BHSig-H and BHSig-B Datasets.}
  \centering
  \begin{tabular}{l||rrrr|rrrr}
    \toprule
    \multirow{2}{*}{Method} & \multicolumn{4}{c|}{BHSig-H} & \multicolumn{4}{c}{BHSig-B} \\
     & ~FAR & ~FRR & ~Acc $\uparrow$ & ~EER $\downarrow$ & ~FAR & ~FRR & ~Acc $\uparrow$ & ~EER $\downarrow$ \\
    \midrule
    SET~\cite{Ren2022SETAS}       & 8.93 & 10.94 & 90.06 & 9.32 & 5.67 & 10.83 & 91.79 & 8.21 \\
    SWIS~\cite{SWIS}              & 10.40 & 59.80 & 72.43 &     - & 36.70 & 11.60 & 72.04 &    -  \\
    \textbf{Ours}                 &  \textbf{1.07} &  \textbf{3.59} & \textbf{98.24} &  \textbf{2.07} &  \textbf{0.95} &  \textbf{4.04} & \textbf{98.19} & \textbf{2.11} \\
    \bottomrule
  \end{tabular}
  \label{tab:table_BHSig_B_H_more}
\end{table}

\begin{table}
\caption{More OSV comparison on CEDAR Dataset.}
  \centering
  \begin{tabular}{l||rrrr}
    \toprule
    Method & ~FAR & ~FRR & ~Acc $\uparrow$ & ~EER $\downarrow$ \\
    \midrule
    SigCNN~\cite{Jiang2022ForgeryfreeSV}              & - & - & - & 6.41 \\
    Consensus~\cite{Brimoh2024ConsensusThresholdCF}   & 5.27 & 17.45 & 92.29 & 11.36 \\
    \textbf{Ours}              & \textbf{0.36} & \textbf{0.58} & \textbf{99.53} & \textbf{0.58} \\
    \bottomrule
  \end{tabular}
  \label{tab:table_CEDAR_more}
\end{table}

}

\begin{table}[t]
\caption{Datasets and protocol description. (\(^a\): number of signers. \(^b\): number of signatures per signer)}
\centering
\begin{tabular}{l|l|l|l|l|ll} 
\toprule
\textbf{Datasets} & \textbf{Language} & \textbf{Total\(^a\)} & \textbf{Genuine\(^b\)} & \textbf{Forged\(^b\)} & \multicolumn{1}{l|}{\textbf{Training\(^a\)}} & \textbf{Testing\(^a\)} \\ 
\midrule
CEDAR~\cite{CEDAR} & English & 55 & 24 & 24 & \multicolumn{1}{l|}{50} & 5 \\
BHSig-H~\cite{BHSig} & Hindi & 160 & 24 & 30 & \multicolumn{1}{l|}{100} & 60 \\
BHSig-B~\cite{BHSig} & Bengali & 100 & 24 & 30 & \multicolumn{1}{l|}{50} & 50 \\
ChiSig~\cite{Yan_2022_CVPR} & Chinese & 500 & 3-5 & above 10 & \multicolumn{1}{l|}{80} & 20 \\
GPDS Synthetic~\cite{Ferrer2015StaticSS} & Western & 4000 & 24 & 30 & \multicolumn{1}{l|}{3200} & 800 \\
MCYT-75~\cite{MCYT} & Western & 75 & 15 & 15 & \multicolumn{1}{l|}{-} & - \\
SigComp 2011~\cite{SigComp2011} & Multi & 64 & 23-24 & 8-20 & \multicolumn{1}{l|}{-} & - \\
\toprule
 & \multicolumn{2}{l|}{\textbf{Training pairs}} & \multicolumn{2}{l}{\textbf{Testing pairs}} &  &  \\ 
\cline{2-5}
 & \textbf{pos} & \textbf{neg} & \textbf{pos} & \multicolumn{1}{l}{\textbf{neg}} &  &  \\ 
\midrule
CEDAR~\cite{CEDAR} & 13800 & 28800 & 1380 & \multicolumn{1}{l}{2880} &  &  \\
BHSig-H~\cite{BHSig} & 27600 & 72000 & 16560 & \multicolumn{1}{l}{43200} &  &  \\
BHSig-B~\cite{BHSig} & 13800 & 36000 & 13800 & \multicolumn{1}{l}{36000} &  &  \\
\bottomrule
\end{tabular}
\label{tab:table_more_datasets}
\end{table}

\CCH{
\section{Overview of Major OSV Datasets}

We have compiled detailed information on several prominent OSV datasets in \cref{tab:table_more_datasets}. These datasets have been extensively utilized and referenced by previous OSV methods. While certain datasets may not be currently accessible, we have conducted our experiments using the datasets currently available. Below, we overview these OSV datasets.

\subsection{ChiSig Dataset}

The ChiSig Dataset \cite{Yan_2022_CVPR} represents a new Chinese document offline signature dataset. It consists of 10,242 images that feature 500 distinct signed names. Given the intricate complexity of Chinese characters, this dataset provides an ideal testbed to evaluate the effectiveness of signature verification models. In our study, we used a subset of 100 signed names from the ChiSig dataset for training and testing purposes. Specifically, 80 of these names were used for the training phase, while the remaining 20 were reserved for testing.

\subsection{CEDAR Dataset}

\MC{The CEDAR Dataset \cite{CEDAR} has a preprocessing detail that has been overlooked by some papers, including Longjam et al.~\cite{Longjam}, 2C2S~\cite{REN2023105639} and DeepHSV~\cite{Li:19DeepHSV}. As pointed out by CaC~\cite{Lu:21Cac} and HybridFE~\cite{Xiong2023HybridFE}, the image normalization is not performed during preprocessing, leading to unfair comparison (achieving 100\% accuracy on CEDAR). We show some raw samples in \cref{fig:CEDAR_sample}. The difference in backgrounds between genuine and forged samples is clearly visible.}

\begin{figure}
    \centering
    \begin{subfigure}{0.4\columnwidth}
        \includegraphics[width=\textwidth]{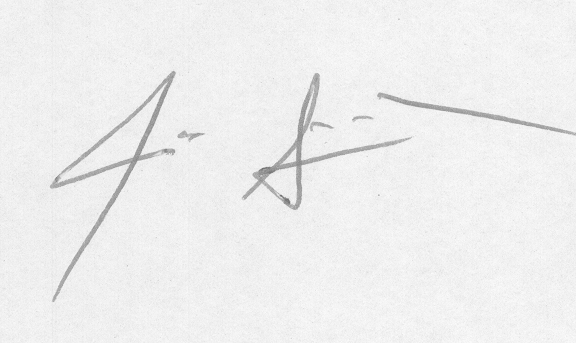}
    \end{subfigure}
    \begin{subfigure}{0.4\columnwidth}
        \includegraphics[width=\textwidth]{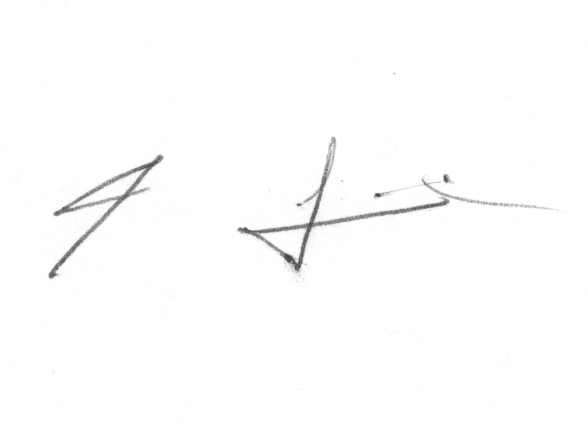}
    \end{subfigure}
    \caption{A genuine (left) and a forged (right) sample of CEDAR. Without image normalization, differences in the background can make the model easily perform verification based on the background rather than the strokes of the signature.}
    \label{fig:CEDAR_sample}
\end{figure}

\subsection{GPDS Synthetic Dataset}

\MC{
The GPDS Synthetic Dataset~\cite{Ferrer2015StaticSS} is a synthetic dataset comprising a substantial volume of image data. Consequently, training models on it requires a significant investment of time and computational resources. Moreover, despite its size, the synthetic nature of the dataset introduces a domain gap between the synthetic and real-world cases. Consequently, only a limited number of studies have utilized it for experimentation. Additionally, this dataset has several predecessors, such as GPDS-960 and GPDS-300, which are no longer accessible.

\subsection{MCYT-75 Dataset}

The MCYT-75 Dataset~\cite{MCYT} is a widely recognized dataset within the OSV domain. It has served as a crucial resource in various prior studies, including those by Maergner et al.\cite{MAERGNER2019527}, Viana et al.\cite{VIANA2023119589}, and Qian Wan and Qin Zou~\cite{Wan2021LearningMF}. Unfortunately, public access to the dataset is currently unavailable. Therefore, seeking assistance from the research groups associated with previous studies becomes necessary.

\subsection{SigComp 2011 Dataset}

The SigComp 2011 Dataset~\cite{SigComp2011} is one of the few OSV datasets that contain multilingual data. It includes signature data in both Dutch and Chinese. One of its drawbacks is the limited amount of data that it provides. The Dutch dataset, for instance, consists of only 362 signatures for training and 1932 signatures for testing, while the Chinese dataset includes 575 signatures for training and 602 signatures for testing. Additionally, its protocol differs from ours as it uses 12 signatures as references, unlike our method, which employs only one signature as a reference. For these reasons, recent methods tend to use other datasets mentioned earlier rather than the SigComp 2011 Dataset for experimental comparisons. Thus, our method is also less straightforward to compare directly with results obtained using this dataset.
}

\section{Preprocessing}

In the preprocessing stage of signature images, we aim to apply normalization and binarization techniques to render them more suitable for input into our model. The process begins with applying a Gaussian filter to the input image, which helps eliminate minor background noise. Subsequently, we employ Otsu's algorithm~\cite{otsu} for image binarization. This step involves finding the center of mass of the image and determining an appropriate threshold to remove background noise effectively.
Once the center of mass is identified, it is used to centrally align the signature within the image. The final step in the preprocessing procedure involves resizing the image to a specific dimension. For our purposes, we resize the image to a standard size of \(224 \times 224\) pixels. This standardized sizing ensures uniformity and consistency in the input data for our model.


\section{Revisiting Earth Mover's Distance (EMD)}

In our study, we utilize the Earth Mover's Distance (EMD) \cite{zhang:22} as an auxiliary metric to evaluate the dissimilarity between two signatures, focusing on their local structure. Let \(\mathcal{R} = \{(r_0, w_{r_0}), (r_1, w_{r_1}), ..., (r_{N-1}, w_{r_{N-1}})\}\) includes \(N\) pair data, where \(r_i\) denotes the \(i_{th}\) feature token of the \(i_{th}\) image patch in the reference signature, with \(w_{r_i}\) representing its corresponding weight. Similarly, \(\mathcal{Q} = \{(q_0, w_{q_0}), (q_1, w_{q_1}), ..., (q_{M-1}, w_{q_{M-1}})\}\), having \(M\) pair data, signifies the pairs of feature tokens and their weights in the Query signature. These feature tokens are derived from a \emph{Reference} image and a \emph{Query} image, respectively, based on our DetailSemNet model. It is important to note that the weights assigned to these tokens can be pre-defined.

To measure the Earth Mover's Distance (EMD) between the two token sets \(\mathcal{R}\) and \(\mathcal{Q}\), we first define the \emph{flow} between them as a matrix \(\emph{F} = [f_{ij}] \in \mathbb{R}^{N \times M}\). Intuitively, \(f_{ij}\) means the amount of importance of the \(i_{th}\) token in \emph{Rererence} ``flowing'' to the \(j_{th}\) token in \emph{Query}. Therefore, \(f_{ij}\) implicitly indicates the confidence level of local token matching.
Next, we employ \(\emph{D} = (d_{ij}) \in \mathbb{R}^{N \times M}\) to denote the ground distance matrix. The distance element \(d_{ij}\) between \((r_i, q_j)\) is calculated as defined in Eq. (6). This equation measures the cosine distance between pairs of token embeddings.

With the matrices (\emph{F}) and (\emph{D}), our objective is to identify the optimal flow \(\emph{F}^*\) that minimizes the constrained objective function outlined in \cref{eqn:cost_function}. This involves calculating the total weighted distances (i.e., \(d_{ij}\)\(f_{ij}\)) between pairs of patch feature tokens from the two sets.



\begin{equation}
COST(\mathcal{R}, \mathcal{Q}, \emph{F}) = \sum_{i=0}^{N-1}\sum_{j=0}^{M-1}d_{ij}f_{ij}
\label{eqn:cost_function}
\end{equation}
subject to the following constraints:
\begin{equation}
f_{ij} \geq 0, 0\leq i\leq N-1, 0\leq j\leq M-1
\end{equation}
\begin{equation}
\sum_{i=0}^{N-1}f_{ij} \leq w_{q_j}, 0\leq i\leq N-1
\end{equation}
\begin{equation}
\sum_{j=0}^{M-1}f_{ij} \leq w_{r_i}, 0\leq j\leq M-1
\end{equation}
\begin{equation}
\sum_{i=0}^{N-1}\sum_{j=0}^{M-1}f_{ij} = \min (\sum_{i=0}^{N-1}w_{r_i} , \sum_{j=0}^{M-1}w_{q_j})
\label{eqn:total_flow}
\end{equation}

After finding the optimal flow \(\emph{F}^*\), the Earth Mover's Distance is determined by the following operation:
\begin{equation}
EMD(\mathcal{R}, \mathcal{Q}) = \frac{\sum_{i=0}^{N-1}\sum_{j=0}^{M-1}d_{ij}f^*_{ij}}{\sum_{i=0}^{N-1}\sum_{j=0}^{M-1}f^*_{ij}}
\label{eqn:emd_}
\end{equation}


It is crucial to recognize that the Earth Mover's Distance (EMD) functions effectively as a metric only when two key conditions are met: first, the two distributions under comparison must have the same total weights, and second, the ground distance function must satisfy metric properties, as outlined in \cite{Rubner:00}. In our work, we assign uniform weights to each feature token and implement weight normalization by \cref{eqn:w_r} and \cref{eqn:w_q} to satisfy these requirements. This approach ensures that the cumulative weight of all features amounts to 1. 

\begin{equation}
w_{r_i} = \frac{1}{N}, 0\leq i\leq N-1
\label{eqn:w_r}
\end{equation}
\begin{equation}
w_{q_j} = \frac{1}{M}, 0\leq j\leq M-1
\label{eqn:w_q}
\end{equation}

Finally, to effectively address the optimization problem outlined in \cref{eqn:cost_function}, we employ the Sinkhorn algorithm \cite{sinkhorn}. This algorithm enhances the traditional Earth Mover's Distance (EMD) by incorporating entropic regularization, which introduces a smoothing term. The inclusion of this term allows the Sinkhorn algorithm to provide an efficient and effective solution to the EMD optimization problem, balancing computational efficiency with the precision required for our analysis.

\begin{table}
  \centering
  \caption{Comparing different \(\lambda_0\) on BHSig-H dataset.}
  \begin{tabular}{l||rrr}
    \toprule
    \(\lambda_0\) & EER(\%) $\downarrow$ & Acc(\%) $\uparrow$ & AUC $\uparrow$ \\
    \midrule
    0.1    & 2.52 & 97.98 & 0.997 \\
    \textbf{1.0}    & \textbf{2.07} & \textbf{98.24} & \textbf{0.998} \\
    10.    & 2.77 & 97.59 & 0.997 \\
    100.   & 3.47 & 96.99 & 0.995 \\
    \bottomrule
  \end{tabular}
  \label{tab:table_lambda_0}
\end{table}
}
\CCH{
\section{The Fusion Ratio between Global Distance and Local Structural Distance}

Our model is specifically designed to calculate the ``distance'' between two input images, a critical factor in determining the final verification result. A key aspect of this process is the appropriate adjustment of the fusion ratio between the Global Distance and the Local Structural Distance. As outlined in Eq. (2), the hyperparameter \(\lambda_0\) is instrumental in balancing the contributions of these two types of distances. Fine-tuning \(\lambda_0\) directly influences the model's overall performance. The experimental findings regarding the optimization of \(\lambda_0\) are comprehensively detailed in \cref{tab:table_lambda_0}. Our experiments indicate that setting \(\lambda_0\) to 1.0 yields the best results in terms of both accuracy and reliability. Therefore, this setting (\(\lambda_0\)=1.0) has been consistently applied across all our experimental configurations.

\begin{table}
  \centering
  \caption{Comparing Various Training Strategies: Our training involves using the BHSig-H dataset and testing on BHSig-B. In this context, the term ``upper bound'' denotes training the model on BHSig-B and testing it on the same dataset, serving as a reference point.}
  \begin{tabular}{l||r}
    \toprule
    Setting & EER(\%) $\downarrow$ \\
    \midrule
    zero-shot    & 7.46 \\
    finetuning   & 4.70 \\
    upper bound & 2.11 \\
    \bottomrule
  \end{tabular}
  \label{tab:table_finetune}
\end{table}

\begin{table}
    \centering
    \caption{Ablation study of w/wo SemanticsAttend on EER (\%). The performance significantly drops without the attention module.}
    \begin{tabular}{c||r|r|r|r}
        \toprule
         & BHSig-H & BHSig-B & CEDAR & ChiSig \\
        \midrule
        $\times$   & 21.06 & 13.29 & 22.14 & 29.01 \\
        \checkmark & \textbf{2.07} & \textbf{2.11} & \textbf{0.58} & \textbf{5.85} \\
        \bottomrule
    \end{tabular}
    \label{tab:table_attention}
\end{table}

\begin{figure}[t]
\centering
\includegraphics[width=0.9\textwidth]{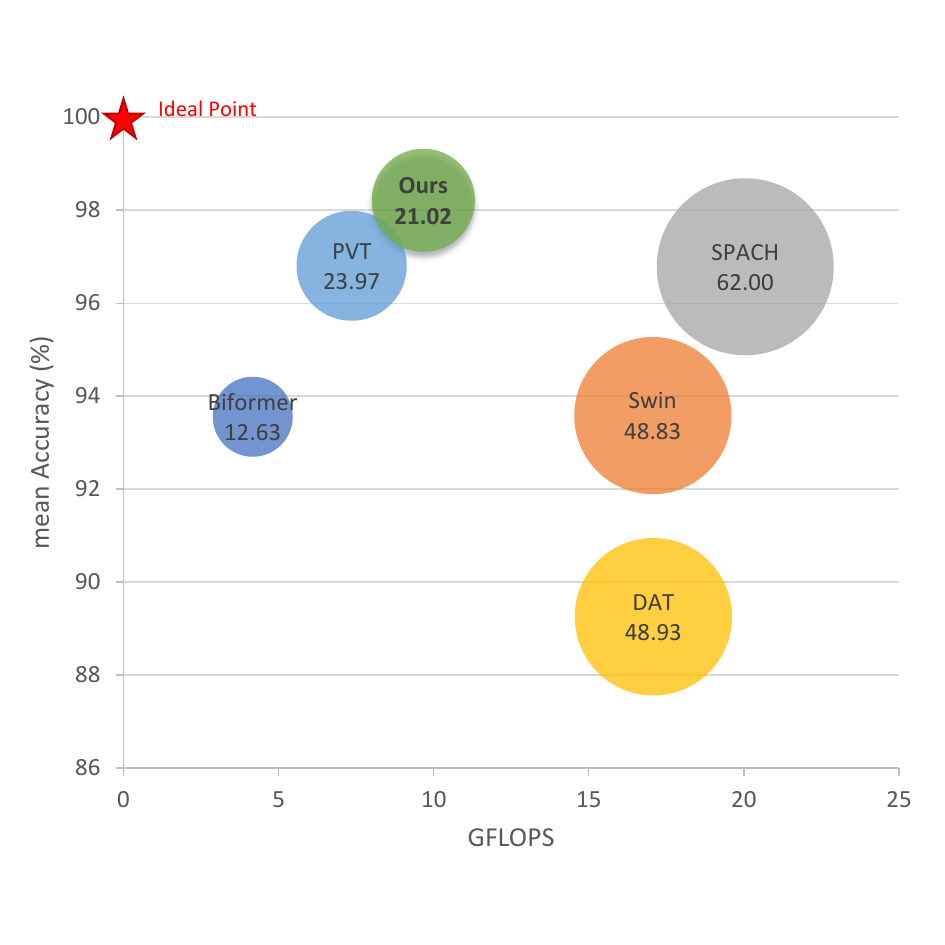}
\caption{Compare our method with different transformer-based architecture. The X-axis represents computational complexity, the Y-axis represents verification accuracy, and the size of the bubbles represents model size (M).}
\label{fig_gflops}
\end{figure}

\section{Cross-dataset Evaluation with Different Training Strategies}

In the context of cross-dataset scenarios, in addition to assessing the model's robustness in a zero-shot setting, we also explore the effectiveness of fine-tuning our model for performance evaluation on different datasets. Fine-tuning enables the model to adapt to specific characteristics of a new dataset, which can lead to improved performance, especially in cases where the training and testing datasets vary significantly.

As detailed in \cref{tab:table_finetune}, we present the results obtained using different training strategies. Here, the ``zero-shot'' testing approach, which is consistent with the methodology described in the sections (Tab. 3) of the main paper, involves training the model on the BHSig-H dataset and subsequently evaluating its performance on the BHSig-B dataset directly without any additional fine-tuning. As for the term ``upper bound'' mentioned in \cref{tab:table_finetune}, it represents the performance achieved in the specific experiment (Tab. 3), where both training and testing occur on the same dataset (BHSig-B in this instance). For the ``fine-tuning'' process, we unfreeze only the parameters of \(Conv_{fuse}\), meaning that the adjustments involve only a limited number of parameters.
The results in \cref{tab:table_finetune}  demonstrate that even with minimal parameter adjustments, the model's performance can closely approach the upper bound. This outcome underscores the model's strong generalization capabilities, suggesting its potential effectiveness across different dataset scenarios.
}

\section{More Ablation Studies on the SemanticsAttend Branch}

Account for the impact of the SemanticsAttend Branch. In terms of results with and without the SemanticsAttend Branch, we present them in \cref{tab:table_attention}. This highlights the effectiveness of our method.

In order to further compare the possibility of using CNNs as backbones in our method. In experiments using CNNs as the backbone, including SM in CNN contributes to worse performance. In the case of using ResNet50~\cite{He2015ResNet} as the backbone and evaluated on the ChiSig dataset, Acc dropped from 87.79\% to 87.28\%. In summary, CNN backbone performs less effectively on OSV than ViT.

\section{GFLOPs and Model Size}

According to Section 4.5, we compared the EER and Acc between various transformer architectures with our method, and here we further consider model size and computational complexity. If our model is to be further used in real-world scenarios, these indicators remain equally important. Detailed results can be seen from \cref{fig_gflops}, where our method achieves the highest accuracy (Y-axis) while maintaining relatively good performance in terms of model parameter size (bubble size, M) and computational complexity (X-axis). From this experiment, we can further observe that our model not only outperforms other methods in verification accuracy, but also exhibits good computational efficiency. This is evidently crucial for real-world deployment scenarios.

\section{Add Classifier}
\label{sec:classifier}

In the \textit{synthetic dataset}~\cite{Ferrer2015StaticSS}, we can additionally integrate a classifier directly into our model, allowing it to learn the common genuine/ forged differences between inter-classes. Specifically, we concatenate two features from the third stage of the siamese backbone. Using a multi-layer convolutional classifier, the final classifier outputs two logits, which are supervised through cross-entropy loss.

To train the classifier part, we employ a cross-entropy loss, defined as follows:
\begin{equation}
p_i = \frac{e^{z_{i}}}{e^{z_0} + e^{z_1}} \ \ \ for\ i=0,1
\end{equation}
\begin{equation}
Loss_{CE} = y log(p_0) + (1 - y)log(p_1).
\end{equation}
Here, the supervised label \(y\) is assigned a value of 1 for positive pairs of signature (genuine-genuine) and 0 for negative (genuine-forged) signature pairs. \(p_0\) and \(p_1\) represent the two logits, \(z_0\) and \(z_1\), output of the classifier after softmax.

In the model, which integrates a classifier, training is performed using \(Loss_{DM}\) combined with \(Loss_{CE}\), defined as
\begin{equation}
Loss_{Total} = Loss_{DM} + Loss_{CE}.
\end{equation}
During inference time, we introduce the CE distance, \(dis_{CE} = z_0 - z_1\). The total distance, if the model includes a classifier, is calculated as
\begin{equation}
dis' = dis + dis_{CE}.
\label{eqn:dis_cls}
\end{equation}

\begin{table}[htp]
\caption{Signature verification comparison on GPDS Synthetic Dataset (\%).}
  \centering
  \begin{tabular}{l||rrrr}
    \hline
    Method & ~FAR & ~FRR & ~Acc $\uparrow$ & ~EER $\downarrow$ \\
    \hline
    SigNet~\cite{dey2017signet}         & 22.24 & 22.24 & 77.76 & 22.24 \\
    Correlated Features~\cite{Dutta:16} & 28.34 & 27.62 & 88.79 & - \\
    SDINet~\cite{Li:21SDINet}           & 12.37 &  8.32 & 89.66 & 10.43 \\
    AVN~\cite{Li:22AVN}                 & 11.78 &  7.58 & 90.32 & 9.77 \\
    TransOSV~\cite{TransOSV_PR}         & 10.64 & 10.64 & - & 10.64 \\
    \hline
    \textbf{Ours} \textit{\#ref=1}      & 11.56 & 11.81 & 88.33 & 11.69 \\
    \textbf{Ours} \textit{\#ref=2}      &  8.60 &  9.81 & 90.89 &  9.15 \\
    \textbf{Ours} \textit{\#ref=3}      &  7.96 &  8.20 & 91.95 &  8.06 \\
    \hline
    \textbf{Ours} \textit{classifier}   &  9.17 &  8.59 & 91.12 &  8.92 \\
    \hline
  \end{tabular}
  \label{tab:table_GPDS}
\end{table}

\section{Results on GPDS Synthetic Dataset}

The GPDS Synthetic Dataset~\cite{Ferrer2015StaticSS} is a synthetic dataset comprising a substantial volume of image data. The previous methods used 3200 signers for training and the remaining 800 signers for testing. Consequently, training models on it requires a significant investment of time and computational resources. Moreover, despite its size, the synthetic nature of the dataset introduces a domain gap between the synthetic and real-world cases. Our model focuses on detecting subtle differences between signature pairs, making it less suitable for the synthetic dataset. However, with minor adjustments to the testing methods or model architecture, our model can still outperform the state-of-the-art methods on the synthetic dataset. First, we can show the testing strategy to use only one reference pair (\textit{\#ref=1}), resulting in an EER of 11.69. By increasing the number of reference pairs, we can mitigate synthetic bias and improve verification performance. Using two reference pairs (\textit{\#ref=2}), the EER is 9.15, and with three reference pairs (\textit{\#ref=3}), the EER is 8.06, surpassing the previous methods, such as SigNet~\cite{dey2017signet}, Correlated Features~\cite{Dutta:16}, SDINet~\cite{Li:21SDINet}, AVN~\cite{Li:22AVN}, and TransOSV~\cite{TransOSV_PR}. We can also enhance our original model with a simple classification architecture, allowing it to directly learn the authenticity features of the synthetic data. With this modification, our model achieves an EER of 8.92, surpassing the performance of the previous best methods. Our method can outperform by using multiple pairs or simply adding a classifier while testing. The results of these comparative methods on the GPDS Synthetic Dataset are detailed in \cref{tab:table_GPDS}. In the following, we will describe the details of our two minor adjustments.

\subsection{Multiple References}

In the GPDS dataset~\cite{Ferrer2015StaticSS}, we improve verification accuracy by increasing the number of reference images. Specifically, each class in the GPDS dataset contains 24 genuine signatures and 30 forged signatures. From the genuine signatures, a specific number (1, 2, or 3) of samples are selected as reference images, and the remaining samples are used as test images. In other words, when the number of reference images is set to 1, 2, and 3, each class contains 23, 44, and 63 positive pairs and 30, 60, and 90 negative pairs, respectively. In this setup, we average the distances obtained from multiple pairs for evaluation.

\subsection{Add Classifier}

In \cref{sec:classifier}, we describe how a simple classifier can be used to improve the accuracy of our model on synthetic datasets. Here, we perform the evaluation using \cref{eqn:dis_cls}. The experimental results are shown in the last row of \cref{tab:table_GPDS}, labeled \textbf{Ours} \textit{classifier}, demonstrating that it surpasses the precision of previous methods. Achieving an accuracy (Acc) of 91.12\%, a False Acceptance Rate (FAR) of 9.17\%, a False Rejection Rate (FRR) of 8.59\%, and an Equal Error Rate (EER) of 8.92\%.


%
%
\bibliographystyle{splncs04}
\bibliography{main}